\documentclass[letterpaper, 10 pt, conference]{ieeeconf}  

\IEEEoverridecommandlockouts                              

\usepackage{booktabs}

\usepackage{graphics} 
\usepackage{epsfig} 
\usepackage[dvipsnames]{xcolor}
\usepackage{amsmath} 
\usepackage{amssymb}  
\usepackage{hyperref} 

\title{\LARGE \bf Open-Source Transtibial Prosthetic Bypass Preserving Joint Alignment: Gait Learning and Kinematic Evaluation
}

\author{Apolline de Vaulchier$^{1}$, Pablo Nocito$^{1}$, Yann Corbel$^{2}$, Nathanaël Jarrassé$^{1}$
\thanks{$^{1}$Sorbonne Université, CNRS, INSERM, Institut des Systèmes Intelligents et de Robotique, ISIR, 75005, Paris, France
        {\tt\small \{devaulchier, nocito, jarrasse\}@isir.upmc.fr}}%
\thanks{$^{2}$Prosthetist-orthotist, Ker Ortho, Rennes, France}%
}

\begin{document}

\maketitle
\thispagestyle{empty}
\pagestyle{empty}

\begin{abstract}
Prosthetic bypasses allow able-bodied individuals to walk with a prosthesis, enabling safe early-stage testing of new prosthetic devices before clinical trials. However, existing transtibial bypasses impose constraints, such as a contralateral shoe lift or a lateral offset of the prosthesis, and their effects on gait remain poorly documented. We present an open-source transtibial prosthetic bypass that preserves part of the sagittal knee mobility without these constraints. Six able-bodied participants familiarized themselves with the bypass for approximately 10~min, then walked on a treadmill with and without it, and joint kinematics were recorded with motion capture and then extracted with OpenSim. All participants were able to walk without handrail support, although single stance remained shorter on the bypass side (27\% vs.\ 45\% of the gait cycle). Bypass-side ankle kinematics were consistent with those reported for transtibial prosthesis users, whereas peak knee flexion during swing decreased from 61$\boldsymbol{^\circ}$ to 36$^\circ$ ($\boldsymbol{p<0.001}$) and hip range of the sound leg increased (35$\boldsymbol{^\circ}$ to 51$^\circ$, $\boldsymbol{p<0.01}$). Although gait with the bypass does not fully replicate that of prosthesis users, it yielded consistent gait patterns after only 10~min of learning. This bypass therefore provides a useful tool for the lower-limb prosthetics research community, facilitating preliminary experiments with able-bodied participants.
\end{abstract}


\section{INTRODUCTION}
Lower-limb amputation affects a large and growing number of people worldwide. According to a study conducted in 2022 by Bruyant et al.~\cite{Bruyant2023}, 116,866 amputations were recorded in France between 2011 and 2020, and in the United States, the number of people living with limb loss could more than double by 2050, mainly due to diabetes and vascular disease~\cite{Ziegler-Graham2008}. Following amputation, several options are available to restore mobility. One of them is wearing a prosthesis: prosthetic devices help restore mobility, and thereby independence and social participation~\cite{Kobayashi2024}.

The research and development of prostheses, and in particular of active prostheses whose control laws must be tuned and validated, relies on extensive experimental testing. Performing these tests directly with people with amputation is demanding: the number of available participants is limited, each trial places an additional burden and risk on them, and ethical and medical approval processes are lengthy. To facilitate these early stages, prosthetic bypass systems have been developed~\cite{Neelen2026}. They allow able-bodied individuals to walk with a prosthesis and are mainly used for prototyping.

Transtibial bypasses fall into two categories. In the first, the prosthesis is attached under the participant's foot by means of a rigid boot fitted with a pyramid adapter~\cite{Louessard2022}. This configuration requires a shoe lift on the contralateral side to compensate for the leg-length difference, and places the prosthetic joint below the anatomical one. In the second, the prosthesis is attached to the side of the leg~\cite{Posh2023}, which avoids an excessive shoe lift but offsets the prosthesis laterally from the leg axis. To our knowledge, no transtibial bypass avoids both the shoe lift and the joint offset without locking the knee. Moreover, the effect of the bypass on gait has mainly been evaluated for transfemoral simulators, whereas such evaluations are lacking for transtibial simulators~\cite{Neelen2026}.

In this context, we propose an open-source transtibial bypass that is simple to build and preserves part of the sagittal knee mobility without requiring a shoe lift. We evaluate the learning of gait with this bypass in able-bodied participants, and then its effect on the sagittal kinematics of the hip, knee and ankle of both limbs, in order to characterise the influence of the bypass on gait.

\section{MATERIAL AND METHODS}
\subsection{PROSTHETIC BYPASS}
\begin{figure}
    \centering
    \includegraphics[width=0.96\linewidth]{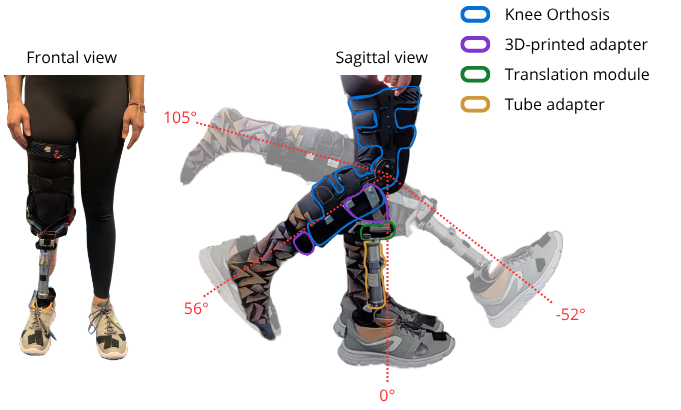}
    \caption{Prosthetic bypass system fitted to an able-bodied participant, with extreme positions shown as transparent overlays. At a maximum knee flexion of 105° in the intact leg on which the bypass is mounted, the prosthetic leg reaches a maximum flexion of 56°. The prosthetic leg also allows a non-physiological extension of 52°.}
    \label{fig:bypass}
    \vspace{-5mm}
\end{figure}

As shown in Figure~\ref{fig:bypass}, the prosthetic bypass system consists of a SecuTec Genu knee orthosis (Bauerfeind) and a 3D-printed adapter connecting the orthosis to the prosthetic components. To allow alignment adjustment, a Janton C61-1 translation module is mounted proximal to the tubular segment corresponding to the shank. This segment consists of an adjustable 2R45=34 tube adapter (Ottobock), which provides angular adjustment of about $\pm14$° and a height adjustment of 6~cm. A strap system secures the prosthetic bypass system to the participant's leg. All parts (including the STEP files of the 3D-printed parts) and detailed assembly instructions are available in the following Git repository (GPL license): \textit{\href{https://gitlab.isir.upmc.fr/devaulchier/open-source-transtibial-prosthetic-bypass}{https://gitlab.isir.upmc.fr/devaulchier/open-source-transtibial-prosthetic-bypass}}.

Any prosthetic component fitted with a pyramid adapter can be attached to the distal end of the shank tube. To date, the following devices have been tested with the bypass: the Össur Balance Foot and Pro-Flex feet, as well as the Open Source Leg, in a version adapted by our laboratory from the mechanical design of version 1 developed at the University of Michigan~\cite{Azocar2018}.

The total mass of the prosthetic bypass system is 2.2~kg (orthosis + 3D printed parts + tubular parts, no foot), for a cost of approximately 1000~€ (materials only).

A video showing walking with the prosthetic bypass system can be seen \href{https://gitlab.isir.upmc.fr/devaulchier/open-source-transtibial-prosthetic-bypass/-/blob/main/media/bypass.mp4}{here}.

\subsection{PARTICIPANTS}
Participants were 6 healthy volunteers between the ages of 22 and 40 years (sex: 1 (F) / 5 (M), age: $28\pm6$ years, body mass: $67\pm9$ kg, height: $178\pm6$ cm). All participants were right-leg dominant.

Participants were free of any medical conditions that could interfere with the protocol, particularly:
conditions involving a disorder of the vestibular system; conditions
limiting the use of the upper or lower limbs and the back; or difficulties performing
moderate physical effort continuously for ten to fifteen minutes. All participants provided informed consent. This experiment was approved by the local ethics committee (CER Sorbonne University) and was conducted in accordance with the Declaration of Helsinki. All participants signed a consent form authorizing their participation in the experiment, the collection of data, and the recording and use of images and videos.

\subsection{EXPERIMENTAL PROTOCOL}
The experimental protocol consists of two parts: a \textit{learning} part, designed to assess the participants' ability to master the prosthetic bypass system, and a \textit{performance} part, designed to evaluate the performance of the prosthetic bypass system. Participants wear shoes fitted with OpenGo connected insoles (Moticon), which measure the plantar pressure under each foot, as well as reflective markers placed on the lower body (see Figure~\ref{fig:bypass}). The OptiTrack motion capture system is used during some phases to record the participant's kinematics. Finally, a stopwatch is used to measure the duration of each learning phase.

The learning part is divided into five phases, during which the participant wears the prosthetic bypass system fitted with the Össur Balance Foot S (K1-K2 class foot):
\begin{itemize}
    \item \textbf{Phase 1 -- Awareness of load distribution:} using visual feedback of the load distribution between both legs, the participant becomes aware of this distribution and of the associated sensation in the leg wearing the prosthetic bypass system. The participant decides when this phase ends, once they feel ready to move on to the next one.
    \item \textbf{Phase 2 -- Recovering a balanced distribution:} without visual feedback, the participant takes small steps in place for approximately 5~s, then returns to a bipedal stance. Once they believe they have recovered a balanced distribution, the visual feedback is restored. The trial is validated if the load difference between the two legs is less than 5~\%. The participant must complete three successful trials to move on to the next phase.
    \item \textbf{Phase 3 -- Gaining confidence in the prosthetic bypass system:} the participant must maintain a single-leg stance on the leg fitted with the prosthetic bypass system for at least 3~s. They must succeed three times to move on to the next phase. The stopwatch is started and stopped visually by the experimenter, when the contralateral foot leaves the ground and when it touches the ground again, respectively.
    \item \textbf{Phase 4 -- First steps:} the participant takes their first steps in a corridor at a self-selected speed, first with the assistance of another person, then unassisted. The phase ends when the participant feels able to walk on the treadmill.
    \item \textbf{Phase 5 -- Treadmill walking:} the participant performs a first 15~s walk on the treadmill at 3~km/h, during which insole and OptiTrack data are recorded. During this first walk, referred to as the \textit{FW}, the participant is allowed to hold on to the treadmill frame, to start with their feet off the belt, and to stop whenever they wish. Recording is then stopped, and the participant may practice freely, performing as many 15~s trials as they wish. Finally, a last 15~s walk, referred to as the \textit{LW}, is recorded (insoles and OptiTrack). This time, the participant must start directly on the belt, must not touch the metal frame, and must keep walking until the treadmill comes to a complete stop. During all treadmill walks, participants are instructed to look straight ahead.
\end{itemize}

The performance evaluation consists of three 15~s walks on the treadmill at 3~km/h for each of the two conditions: the first with the Balance Foot, and the second with the Össur Pro-Flex foot (K3-K4 class foot). During each walk, the insoles record plantar pressure and the OptiTrack system records the kinematics.

Before the prosthetic bypass system is fitted, three 15~s walks at 3~km/h are recorded to establish the participant's baseline.

At the end of the session, participants answer two questions:
\begin{itemize}
    \item On a scale from 1 to 10 (10 = very comfortable), how comfortable is the prosthetic bypass system?
    \item On a scale from 1 to 10 (10 = very easy), how easy and intuitive is it to learn to use the prosthetic bypass system?
\end{itemize}

\subsection{DATA ANALYSIS}
\paragraph{Joint kinematics}
Gait kinematics were obtained by inverse kinematics in OpenSim (\textit{Inverse Kinematics} tool) from the marker trajectories recorded with the OptiTrack system. The 3DGaitModel2392 model~\cite{Delp1990, Templin2021} was scaled for each participant and each condition (walking without the bypass, walking with the Balance foot and walking with the Pro-Flex foot). Marker errors during inverse kinematics were on average 15~mm (RMS) and at most 35~mm, below the recommended thresholds of 20~mm (RMS) and 40~mm (maximum)~\cite{Hicks2015}. Three trials were recorded per participant and per condition, i.e.\ 18 trials for walking without the bypass and with the Balance foot ($n = 6$). For one participant, the Pro-Flex trials could not be used because of marker occlusions and were excluded, leaving 15 Pro-Flex trials ($n = 5$).

\paragraph{Data processing}
Joint angles were low-pass filtered with a second-order Butterworth filter (cut-off frequency: 5~Hz). Gait was then segmented into cycles based on right heel strikes, detected from the forces measured by the OpenGo pressure insoles. Each right cycle was divided into four phases: double support, right single support, double support and right swing. Left-limb cycles were computed from two consecutive right cycles, from one left heel strike to the next. Several categories of cycles were excluded:
\begin{itemize}
    \item the first two cycles and the last cycle of each trial, to remove the effects of the treadmill starting and stopping;
    \item cycles that did not show the expected phase sequence (double support, single support, double support, swing);
    \item cycles containing a stumble;
    \item cycles whose duration differed by more than 30~\% from the participant's median cycle duration in that condition.
\end{itemize}
After applying these criteria,  135 without the bypass, 128 with the Balance foot and 96 with the Pro-Flex right gait cycles remained (respectively 116, 111 and 84 left gait cycles). Gait variables were computed on these cycles. Each cycle was time-normalised from 0 to 100~\% of the gait cycle (101 points). Angles are positive in ankle dorsiflexion, knee flexion and hip flexion.

\paragraph{Statistical methods}
Learning was assessed on the cycle duration and on the single-support time of the bypass limb and of the sound limb, expressed as a percentage of the gait cycle. Mean values were compared with paired t-tests, and the between-participant dispersion with the Pitman-Morgan test for paired variances. Normality was checked with the Shapiro-Wilk test. When it was not met, the Wilcoxon signed-rank test replaced the t-test, and a rank-based version of the Pitman-Morgan test (Spearman correlation between the sum and the difference of the two conditions) was used ($\alpha = 0.05$).

For the performance analysis: For each joint (ankle, knee, hip), side, participant and condition, the following variables were computed on each cycle and then averaged over cycles:
\begin{itemize}
    \item the range of motion (ROM);
    \item the functional peaks and their time of occurrence as a percentage of the gait cycle;
    \item the angular velocity, obtained by differentiating the angles with respect to time on each cycle before time normalisation, together with its range and its maximum and minimum values and their timing;
\end{itemize}
The ROM of joints and the maximum knee flexion in swing were compared between walking without and with the bypass using a paired Student's t-tests ($\alpha = 0.05$), after checking the normality of the differences with the Shapiro-Wilk test. The other variables are reported descriptively (mean $\pm$ standard deviation across participants). A Holm-Bonferroni correction was applied to the k = 4 comparisons of joint kinematics.

\section{RESULTS}
\begin{table}[htbp]
\centering
\caption{Learning characteristics (mean $\pm$ SD).}
\label{tab:learning}
\begin{tabular}{lc}
\toprule
Parameter & Mean $\pm$ SD \\
\midrule
Training time, phase 1 (min)         & 1.2 $\pm$ 1.2$^{*}$ \\
Training time, phase 2 (min)         & 2.1 $\pm$ 0.5 \\
Training time, phase 3 (min)         & 2.0 $\pm$ 1.1 \\
Training time, phase 4 (min)         & 4.0 $\pm$ 1.2$^{*}$ \\
Number of learning walks, phase 5     & 6 $\pm$ 1$^{*}$ \\
Total learning time (min)            & 9.4 $\pm$ 2.2 \\
\bottomrule
\multicolumn{2}{l} \footnotesize $^{*}$ Duration determined by a participant-dependent criterion.
\end{tabular}
\vspace{-5mm}
\end{table}

\subsection{Learning}
\begin{figure}[ht]
    \centering
    \includegraphics[width=0.96\linewidth]{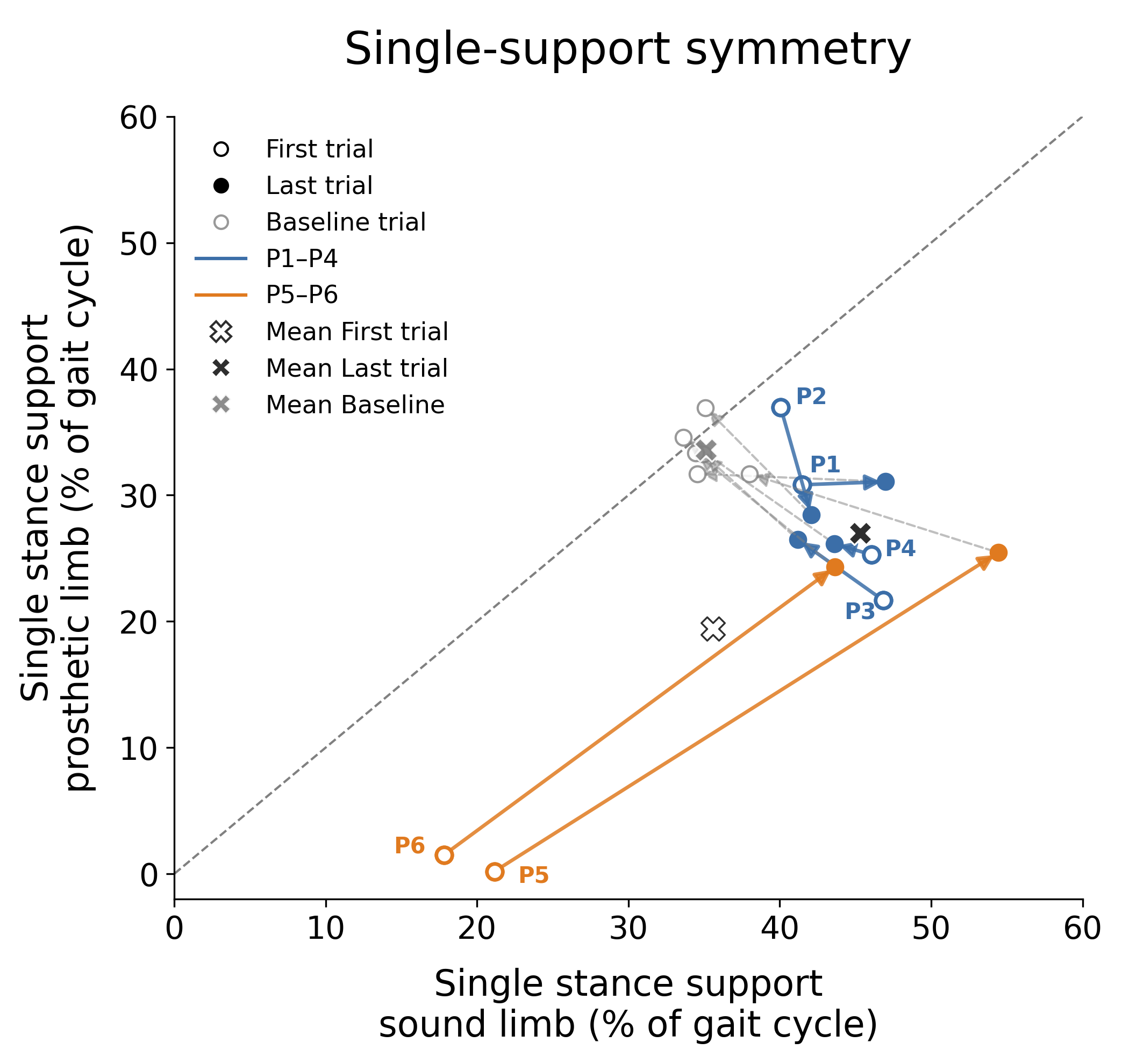}
    \caption{Evolution of single-support symmetry during First, Last and Baseline walking trials for each subject.}
    \label{fig:learning}
    \vspace{-5mm}
\end{figure}

The complete learning session lasted $9.4 \pm 2.2$~min (Table~\ref{tab:learning}), including around 6 training walks of 15s on the treadmill. Between the FW and the LW, temporal parameters moved closer to baseline and their between-participant dispersion decreased markedly (Table~\ref{tab:gait_learning}, Fig.~\ref{fig:learning}). Cycle duration went from $1.1 \pm 0.3$~s to $1.2 \pm 0.1$~s (baseline: $1.3 \pm 0.1$~s), with a significant reduction in dispersion ($p<0.01$). Single support on the bypass limb increased from $19 \pm 15$~\% to $27 \pm 2$~\% of the gait cycle ($p<0.001$ for dispersion), two participants showing almost no single support on this limb during the FW, but remained shorter than at baseline ($34 \pm 2$~\%, $p<0.05$). Conversely, single support on the sound limb was longer than at baseline during the LW ($45 \pm 5$~\% vs.\ $35 \pm 2$~\%, $p<0.001$), so that gait remained asymmetric after learning. At the end of the session, participants rated the comfort of the bypass $6.5 \pm 1.0$ out of 10 and the ease and intuitiveness of learning $6.3 \pm 1.8$ out of 10.

\begin{table}[htbp]
\centering
\caption{Gait parameters for the first, the last with the bypass prosthesis (Balance Foot) and baseline walking (mean $\pm$ SD).}
\label{tab:gait_learning}
\setlength{\tabcolsep}{5pt}
\begin{tabular}{lccc}
\toprule
Parameter & First trial & Last trial & Baseline \\
\midrule
Number of cycles         & 9 $\pm$ 1 & 11 $\pm$ 0 & 11 $\pm$ 1 \\
Cycle duration (s)       & 1.1 $\pm$ 0.3$^{b}$ & 1.2 $\pm$ 0.1 & 1.3 $\pm$ 0.1 \\
Bypass-side single support (\%)     & 19 $\pm$ 15$^{c}$ & 27$^{*}$ $\pm$ 2 & 34 $\pm$ 2 \\
Sound-side single support (\%)     & 36 $\pm$ 13$^{a}$ & 45$^{***}$ $\pm$ 5$^{b}$ & 35 $\pm$ 2 \\
\bottomrule
\multicolumn{4}{l}{\footnotesize $^{*}$p$<$.05, $^{**}$p$<$.01, $^{***}$p$<$.001 for paired T-tests} \\
\multicolumn{4}{l}{\footnotesize $^{a}$p$<$.05, $^{b}$p$<$.01, $^{c}$p$<$.001 for Pitman-Morgan tests} \\
\end{tabular}
\vspace{-4mm}
\end{table}

\subsection{Performance}
\begin{figure}
    \centering
    \includegraphics[width=\linewidth]{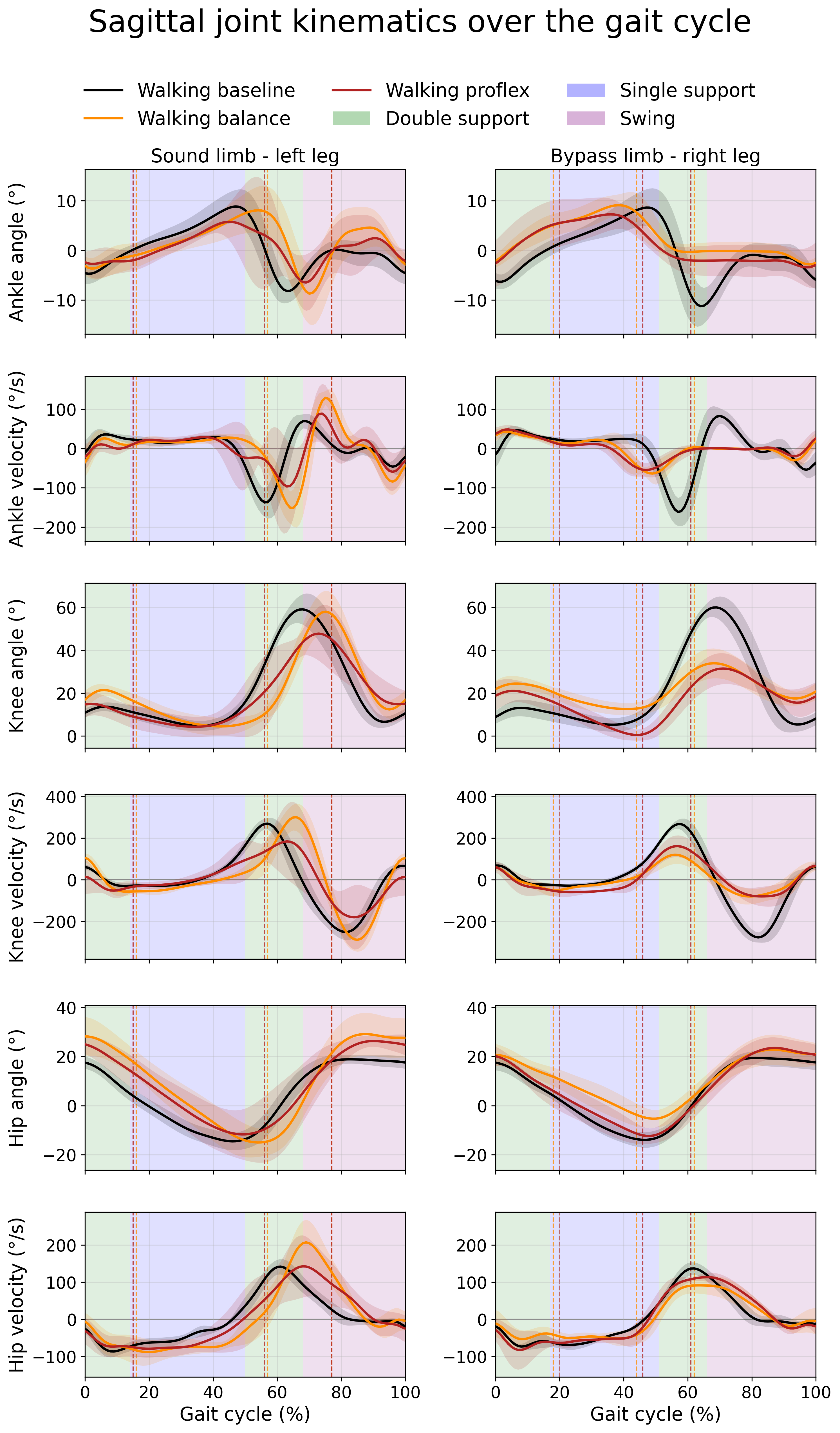}
    \caption{Sagittal angles and angular velocities of the ankle, knee and hip of each leg (mean $\pm$ SD). Variables are expressed as a function of the gait cycle (\%).}
    \label{fig:kinematics}
    \vspace{-5mm}
\end{figure}

Walking with the bypass altered the sagittal kinematics of both lower limbs compared with unimpaired walking (Table~\ref{tab:kinematics}, Fig.~\ref{fig:kinematics}). On the sound limb, most angular peaks were delayed by 4 to 10\% of the gait cycle (GC), consistent with the longer single stance observed on this side after learning.

\paragraph{Ankle}
Maximum dorsiflexion in stance kept its baseline value on both sides (9 vs $10^\circ$), but was reached earlier on the bypass side (47 vs 39\%~GC) and later on the sound side (47 vs 53\%~GC). On the bypass side, push-off plantarflexion almost disappeared ($-12$ vs $-4^\circ$), and its peak moved from push-off to the end of swing (67 vs 92\%~GC), resulting in a significantly smaller ankle ROM (22 vs $14^\circ$, $p=0.024$). On the sound side, ankle ROM (22 vs $24^\circ$, $p=0.80$) and peak plantarflexion ($-12^\circ$ in both conditions) remained similar to baseline. A large asymmetry was observed between the angular velocities of the sound and bypass ankles (range: 412 vs $152^\circ$/s). On the bypass side, the peak dorsiflexion velocity, which occurs in early swing at baseline (65\%~GC), appeared in early stance (22\%~GC).
 
 \paragraph{Knee}
On the bypass side, maximum knee flexion in swing decreased from $61^\circ$ at baseline to $36 \pm 7^\circ$ ($p<0.001$), with a similar timing (68 vs 72\%~GC), and knee ROM was almost halved (59 vs $32^\circ$, $p<0.001$). Knee flexion during loading response increased (14 vs $25^\circ$) and occurred at the same instant (8\%~GC). Maximum knee extension in single stance remained close to baseline (4 vs $6^\circ$, 35 vs 39\%~GC). On the sound side, knee ROM was preserved ($59^\circ$ in both conditions, $p=1.00$), but its pattern changed: flexion during loading response increased (14 vs $21^\circ$), maximum extension in stance occurred later (35 vs 40\%~GC) for a similar value (4 vs $3^\circ$), and maximum flexion in swing tended to decrease (61 vs $56^\circ$, $p=0.27$) and was delayed by 10\%~GC (68 vs 78\%~GC). A large asymmetry was observed between the angular velocities of the sound and bypass knees (range: 705 vs $310^\circ$/s). On the bypass side, the timing of the peak knee extension velocity became highly variable across participants ($62 \pm 22$ vs $82 \pm 2$\%~GC).

\paragraph{Hip}
At the hip, ROM was unchanged ($35^\circ$) for bypass side but significantly increased ($51^\circ$) for sound side ($p=0.004$ vs baseline). For the bypass side the maximum extension in stance was reduced ($-15$ vs $-10^\circ$) and maximum flexion in swing slightly increased (20 vs $24^\circ$), with timings delayed relative to baseline (47 vs 48\% and 84 vs 88\%~GC). On the other side, the maximum extension in stance was increased ($-15$ vs $-18^\circ$) and maximum flexion in swing increased (20 vs $31^\circ$), with timings delayed relative to baseline (47 vs 53\% and 84 vs 91\%~GC). For the bypass side, the hip velocity remained similar to baseline but on the sound side we observed an increase (range: 241 vs $364^\circ$/s; peak flexion velocity: 151 vs $246^\circ$/s).

\paragraph{Balance S vs Pro-Flex feet}
This descriptive comparison includes the 5 participants who walked with both feet. Both feet produced the same overall gait pattern, but several differences were consistent across participants. On the bypass side, the knee was almost fully extended in stance with the Pro-Flex ($0 \pm 3^\circ$), whereas it remained flexed with the Balance ($10 \pm 3^\circ$). The Pro-Flex was also associated with larger knee ROM (35 vs $29^\circ$) and hip ROM (38 vs $35^\circ$), and with greater hip extension in stance ($-14$ vs $-8^\circ$, close to baseline). Ankle ROM was slightly smaller with the Pro-Flex (13 vs $15^\circ$). On the sound side, knee flexion during loading response was smaller with the Pro-Flex (17 vs $23^\circ$).

\begin{table*}[t]
  \centering
  \caption{Sagittal joint kinematics during treadmill walking without and with the bypass (mean $\pm$ SD).}
  \label{tab:kinematics}
  \footnotesize
  \setlength{\tabcolsep}{3pt}
  \begin{tabular*}{\textwidth}{@{\extracolsep{\fill}}l cc cccc@{}}
    \toprule
    & \multicolumn{2}{c}{Baseline} & \multicolumn{4}{c}{Bypass} \\
    \cmidrule(lr){2-3}\cmidrule(l){4-7}
    & \multicolumn{2}{c}{} & \multicolumn{2}{c}{Sound limb (left)} & \multicolumn{2}{c}{Bypass limb (right)} \\
    \cmidrule(lr){4-5}\cmidrule(l){6-7}
    Parameter & Value & \%~GC & Value & \%~GC & Value & \%~GC \\
    \midrule
    \multicolumn{7}{@{}l}{\textbf{Joint angles (deg)}} \\
    \multicolumn{7}{@{}l}{\textit{Ankle}} \\
    \quad ROM & $22 \pm 4$ &  & $24 \pm 4$ &  & $\boldsymbol{14 \pm 1^{*}}$ &  \\
    \quad Max.\ dorsiflexion, stance & $9 \pm 4$ & $47 \pm 4$ & $10 \pm 4$ & $53 \pm 4$ & $10 \pm 3$ & $39 \pm 5$ \\
    \quad Max.\ plantarflexion, push-off & $-12 \pm 2$ & $67 \pm 5$ & $-12 \pm 7$ & $71 \pm 6$ & $\boldsymbol{-4 \pm 2}$ & $92 \pm 10$ \\
    \multicolumn{7}{@{}l}{\textit{Knee}} \\
    \quad ROM & $59 \pm 7$ &  & $59 \pm 4$ &  & $\boldsymbol{32 \pm 3^{***}}$ &  \\
    \quad Flexion, loading response & $14 \pm 3$ & $8 \pm 1$ & $\boldsymbol{21 \pm 4}$ & $5 \pm 2$ & $\boldsymbol{25 \pm 4}$ & $8 \pm 4$ \\
    \quad Max.\ extension, single stance & $4 \pm 3$ & $35 \pm 3$ & $3 \pm 5$ & $40 \pm 6$ & $6 \pm 4$ & $39 \pm 3$ \\
    \quad Max.\ flexion, swing & $61 \pm 7$ & $68 \pm 2$ & $56 \pm 10$ & $78 \pm 3$ & $\boldsymbol{36 \pm 7^{***}}$ & $72 \pm 7$ \\
    \multicolumn{7}{@{}l}{\textit{Hip}} \\
    \quad ROM & $35 \pm 4$ &  & $\boldsymbol{51 \pm 7^{**}}$ &  & $35 \pm 7$ &  \\
    \quad Max.\ extension, stance & $-15 \pm 4$ & $47 \pm 1$ & $-18 \pm 6$ & $53 \pm 6$ & $-10 \pm 4$ & $48 \pm 4$ \\
    \quad Max.\ flexion, swing & $20 \pm 2$ & $84 \pm 5$ & $\boldsymbol{31 \pm 7}$ & $91 \pm 3$ & $24 \pm 5$ & $88 \pm 1$ \\
    \midrule
    \multicolumn{7}{@{}l}{\textbf{Angular velocities (deg/s)}} \\
    \multicolumn{7}{@{}l}{\textit{Ankle}} \\
    \quad Range & $300 \pm 40$ &  & $\boldsymbol{412 \pm 56}$ &  & $\boldsymbol{152 \pm 13}$ &  \\
    \quad Max.\ (dorsiflexion) & $107 \pm 14$ & $65 \pm 3$ & $\boldsymbol{181 \pm 28}$ & $72 \pm 5$ & $\boldsymbol{57 \pm 5}$ & $22 \pm 9$ \\
    \quad Min.\ (plantarflexion) & $-193 \pm 32$ & $57 \pm 2$ & $-231 \pm 41$ & $65 \pm 7$ & $\boldsymbol{-95 \pm 8}$ & $49 \pm 5$ \\
    \multicolumn{7}{@{}l}{\textit{Knee}} \\
    \quad Range & $582 \pm 60$ &  & $\boldsymbol{705 \pm 81}$ &  & $\boldsymbol{310 \pm 32}$ &  \\
    \quad Max.\ (flexion) & $295 \pm 21$ & $57 \pm 2$ & $362 \pm 47$ & $64 \pm 5$ & $\boldsymbol{183 \pm 24}$ & $55 \pm 3$ \\
    \quad Min.\ (extension) & $-286 \pm 42$ & $82 \pm 2$ & $-343 \pm 46$ & $81 \pm 4$ & $\boldsymbol{-127 \pm 20}$ & $62 \pm 22$ \\
    \multicolumn{7}{@{}l}{\textit{Hip}} \\
    \quad Range & $241 \pm 25$ &  & $\boldsymbol{364 \pm 49}$ &  & $232 \pm 58$ &  \\
    \quad Max.\ (flexion) & $151 \pm 17$ & $61 \pm 2$ & $\boldsymbol{246 \pm 41}$ & $67 \pm 5$ & $121 \pm 17$ & $62 \pm 3$ \\
    \quad Min.\ (extension) & $-90 \pm 11$ & $13 \pm 4$ & $\boldsymbol{-118 \pm 13}$ & $23 \pm 9$ & $-111 \pm 44$ & $32 \pm 8$ \\
    \bottomrule
  \end{tabular*}
  \par\vspace{1mm}
  \begin{minipage}{\textwidth}
    \footnotesize
    $^{***}p<.001$, $^{**}p<.01$, paired t-test vs same-side baseline after
    Holm-Bonferroni correction (tests on ROM of joint and knee swing flexion).
  \end{minipage}
\vspace{-4mm}
\end{table*}

\section{DISCUSSION}

We presented an open-source transtibial bypass that allows knee flexion without offsetting joint alignment~\cite{Posh2023} or requiring a shoe lift~\cite{Louessard2022}, two drawbacks of existing simulators~\cite{Neelen2026}. In particular, the absence of a shoe lift preserves plantar feedback on the sound limb. The translation platform, the tube height adjustment and the pyramid adapters allow the bypass to be adapted to each user. This adjustment, however, requires knowledge of prosthetic fitting: although the experimenter had been trained by a prosthetist, it remained long and delicate. This partly explains the moderate ease-of-learning rating ($6.3 \pm 1.8$/10), as some participants included the fitting of the device in their evaluation.

After about 10~min of learning, all participants, with no prior experience, were able to walk on the treadmill with the bypass, including the two participants whose single stance on the bypass limb was almost zero during the first walk. Gait nevertheless remained asymmetric, with more time in single stance on the sound limb (45\% of the gait cycle) than on the bypass limb (27\%). Zhang et al.~\cite{Zhang2019} reported near-perfect symmetry in experienced transtibial amputees, but an asymmetry in the same direction when a new foot was first fitted (38 vs 42\%). This asymmetry therefore seems related to a lack of familiarity, especially as reported learning durations for simulators range from 45~min to 6~months~\cite{Neelen2026}. The greater dispersion between baseline and Last trials is probably due to the baseline walks being recorded at the beginning of the session, before some participants were used to the treadmill.

With the bypass, participants recovered a consistent gait pattern. The increased velocities on the sound side and the stance asymmetry, however, suggest that they did not yet fully trust the device. In experienced transtibial amputees, differences from non-amputees concern joint moments more than joint angles, except at the prosthetic ankle and knee~\cite{Sagawa2011}. Our results differ from this pattern mainly at the knee and the hip.

Compared with the data of Sanderson et al.~\cite{Sanderson1997} in amputees fitted with a Flex-Foot, the bypass ankle closely reproduced the behaviour of a prosthesis: identical range of motion ($14^\circ$ in both studies) and velocity range (150 vs $152^\circ$/s), loss of push-off plantarflexion, and peak dorsiflexion velocity during stance. The knee was the main difference: its range of motion is normal in amputees ($62^\circ$) but was halved with the bypass ($32^\circ$). Since the knee is held flexed in the bypass, part of its natural range is lost, and the short learning period may not have allowed participants to fully develop knee motion. At the hip, compensation occurred in the opposite direction: range of motion increases on the prosthetic side in amputees (35 vs $39^\circ$), but on the sound side this range decreased (51 vs $32^\circ$). Apart from the bypass-side knee, angular velocities were close to those of Lemaire et al.~\cite{Lemaire1993} at the ankle and knee, but higher at the sound hip.

However, the asymmetry observed in our participants is closer to that reported in individuals with recent amputations. In the 12 transtibial amputees followed by Barnett et al.~\cite{Barnett2009} during initial rehabilitation, ankle range of motion was very close to ours (prosthetic: 15 vs $14^\circ$; intact: 22 vs $23^\circ$). Knee range of motion was also smaller on the bypass side ($40^\circ$ vs $32^\circ$ in our study), with peak flexion occurring earlier than on the intact side (80 vs 85\% of the gait cycle, vs 72 and 78\% in our study). Hip range of motion, however, was larger and more asymmetric with the bypass (prosthetic: 30 vs $35^\circ$; intact: 35 vs $51^\circ$), possibly due to a higher walking speed (0.31 to 0.71~m/s in Barnett et al.).

The bypass currently allows walking and static tasks. For more demanding tasks, the leg attachment should be improved, for example with a closed shell tightened by a BOA-type system, which could also improve symmetry and loading comfort by approaching socket conditions~\cite{Board2001}. Another improvement would be to limit the extension of the prosthetic knee to a physiological range, as it currently allows approximately 52° of extension.

This study has several limitations. The bypass was not fitted by a prosthetist. The short familiarization period makes it difficult to separate the effects of the bypass from those of limited training, and longer adaptation might reduce some asymmetries. The imposed treadmill speed may have exceeded participants' spontaneous speed. Marker occlusions limited the number of usable trials, and some markers were repositioned between conditions to accommodate the device, possibly introducing small kinematic offsets. The unmodified gait2392 model does not represent the prosthetic components. The computed ankle angle on the bypass side therefore reflects the kinematic reconstruction rather than the deformation of the prosthetic blade alone. Amputee-specific adaptations~\cite{Templin2021} could address this in future work. The analysis was also restricted to a subset of sagittal-plane variables during level treadmill walking, limiting the comparison with prosthesis users to these conditions. Regarding the device, the flexed knee position prevents full range of motion, precluding tasks such as sitting down or climbing high steps, and the orthosis strongly restricts non-sagittal rotations, making the knee act more like a pivot. Finally, the sample size was small ($n = 6$).

\section{CONCLUSION}

We presented an open-source transtibial prosthetic bypass that preserves part of the sagittal knee mobility without requiring a shoe lift or offsetting joint alignment. Able-bodied participants with no prior experience learned to walk with the device in about 10~minutes, although gait remained asymmetric. The bypass faithfully reproduced the ankle kinematics of transtibial prosthesis users, but knee flexion on the bypass side was reduced and hip range of motion of the sound limb increased. These results suggest that the bypass is an useful tool for the research community on the lower limb prosthetics, allowing an easier way to perform preliminary experiments with asymptomatic participants. Future work will focus on improving the leg attachment and on assessing the effect of longer training on gait symmetry and knee flexion of the bypass side.

\addtolength{\textheight}{-0.5cm}   



\section*{ACKNOWLEDGMENT}

We would like to thank all the participants for their time and involvement in this study.

The authors used Claude in order to improve the language and readability of the manuscript during the translation process. The authors have verified the final text to ensure technical accuracy.


\end{document}